\documentclass[conference]{IEEEtran}
\IEEEoverridecommandlockouts

\usepackage{cite}
\usepackage{amsmath,amssymb,amsfonts}
\usepackage{algorithmic}
\usepackage{url}
\usepackage{graphicx}
\usepackage{svg}
\usepackage{textcomp}
\usepackage{makecell}
\usepackage{enumitem}
\usepackage{diagbox}
\usepackage[hidelinks]{hyperref}
\usepackage{xcolor}
\usepackage{subcaption}
\usepackage{dblfloatfix} 

\def\BibTeX{{\rm B\kern-.05em{\sc i\kern-.025em b}\kern-.08em
    T\kern-.1667em\lower.7ex\hbox{E}\kern-.125emX}}
\newdimen\NetTableWidth
\NetTableWidth=\dimexpr
\textwidth
- 14\tabcolsep
\relax

\usepackage[mathlines,switch]{lineno}

\makeatletter

\let\old@ps@IEEEtitlepagestyle\ps@IEEEtitlepagestyle
\def\confheader#1{%
    \def\ps@IEEEtitlepagestyle{%
        \old@ps@IEEEtitlepagestyle%
        \def\@oddhead{\strut\hfill#1\hfill\strut}%
        \def\@evenhead{\strut\hfill#1\hfill\strut}%
    }%
    \ps@headings%
}
\makeatother

\confheader{%
        {49th IEEE International Conference on Computers, Software, and Applications (COMPSAC 2025) (Accepted version)}
}

\begin{document}

\onecolumn
© 2025 IEEE. Personal use of this material is permitted. Permission from IEEE must be obtained for all other uses, in any current or future media, including reprinting/republishing this material for advertising or promotional purposes, creating new collective works, for resale or redistribution to servers or lists, or reuse of any copyrighted component of this work in other works. 
\twocolumn

\title{Integrating Knowledge Graphs and Bayesian Networks: A Hybrid Approach for Explainable Disease Risk Prediction\\
\thanks{This work is based on research supported in part by the National Research Foundation of South Africa (grant number 151217), the Hasso Plattner Institute for Digital Engineering through the HPI Research School at the University of Cape Town, and the Institute for Humane Studies (grant number IHS019183).}
}

\author{\IEEEauthorblockN{Mbithe Nzomo}
\IEEEauthorblockA{\textit{Centre for Artificial Intelligence Research (CAIR);} \\
\textit{Department of Computer Science} \\
\textit{University of Cape Town} \\
Cape Town, South Africa \\
mnzomo@cs.uct.ac.za}
\and
\IEEEauthorblockN{Deshendran Moodley}
\IEEEauthorblockA{\textit{Centre for Artificial Intelligence Research (CAIR);} \\
\textit{Department of Computer Science} \\
\textit{University of Cape Town} \\
Cape Town, South Africa \\
deshen.moodley@uct.ac.za}
}

\maketitle

\begin{abstract}
Multimodal electronic health record (EHR) data is useful for disease risk prediction based on medical domain knowledge. However, general medical knowledge must be adapted to specific healthcare settings and patient populations to achieve practical clinical use.
Additionally, risk prediction systems must handle uncertainty from incomplete data and non-deterministic health outcomes while remaining explainable. 
These challenges can be alleviated by the integration of knowledge graphs (KGs) and Bayesian networks (BNs). We present a novel approach for constructing BNs from ontology-based KGs and multimodal EHR data for explainable disease risk prediction. 
Through an application use case of atrial fibrillation and real-world EHR data, we demonstrate that the approach balances generalised medical knowledge with patient-specific context, effectively handles uncertainty, is highly explainable, and achieves good predictive performance.
\end{abstract}

\begin{IEEEkeywords}
Knowledge graph, Ontology, Bayesian network, Electronic health records, Disease risk prediction
\end{IEEEkeywords}

\section{Introduction}
\label{intro}
Non-communicable diseases pose a critical threat to global health, claiming 43 million lives each year \cite{who_ncds_2024}. 
AI-based personal health monitoring systems have high potential for predicting and preventing such diseases. 
By identifying at-risk individuals before the onset of disease, health outcomes can be significantly improved with targeted preventive measures. 
This requires the analysis of multimodal health data from electronic health records (EHRs), which presents several challenges. 
A significant one is interoperability. Beyond syntactic and semantic interoperability, risk prediction systems must also support pragmatic interoperability, in which the specific context of the data informs the way it is used \cite{asuncion_pragmatic_2010}. 
While syntactic interoperability ensures compatible data formats and semantic interoperability establishes shared terminology \cite{benson_principles_2021}, neither addresses the adaptation of general medical knowledge to specific clinical contexts. 
Pragmatic interoperability addresses this limitation and sets the foundation for clinical interoperability, which enables different clinical teams to coordinate effectively and seamlessly in patient care based on a shared understanding of patient needs \cite{benson_principles_2021}. 
For health risk prediction, this means that general medical domain knowledge must be adapted and tailored to the particular healthcare setting and patient population where it is used.

Another challenge is that risk prediction is an inherently uncertain endeavour due to incomplete data and the non-deterministic nature of health outcomes. Risk prediction systems must therefore be equipped to handle this uncertainty. 
Finally, explainability has surfaced as a critical challenge in AI systems, with concerns raised about the use of opaque black-box models in health systems \cite{amann_explainability_2020}. 
While the use of inherently explainable models is an important step, this alone is insufficient for health decision support. 
AI systems must go beyond technical explanations of the models and data and provide user-centred explanations that connect to established domain knowledge \cite{chari_explanation_2024}. 
In his theory of explanatory coherence, Thagard \cite{thagard_explanatory_1989} raises alignment with established knowledge as a crucial measure of the acceptability of explanations. 

Two complementary AI techniques can alleviate these challenges: knowledge graphs (KGs) and Bayesian networks (BNs). 
Commonly built using an ontology-based data schema, KGs can formalise domain knowledge and facilitate the querying, reasoning, and visualization of data, enhancing explainability and semantic interoperability \cite{rajabi_knowledge_2022}. 
However, KGs rely on manually curated deductive knowledge \cite{hogan_knowledge_2022}, limiting their adaptability to variations in clinical settings. 
Moreover, KGs do not inherently support uncertainty. 
BNs provide inherent support for reasoning under uncertainty and are used to model probabilistic and causal relationships, providing explainability through causal inference \cite{korb_bayesian_2011}. 
They can be trained on observational data from particular clinical settings while incorporating domain knowledge \cite{mclachlan_bayesian_2020}. 
This hybrid approach combines data-driven methods that capture real-world patterns with knowledge-driven methods that incorporate established medical knowledge.

This paper presents an approach to construct BNs from ontology-based KGs and multimodal EHR data for explainable disease risk prediction. 
We develop an ontology to define the schema of domain knowledge about factors that influence disease risk. 
Building upon the ontology, we develop a KG that stores both the formalised domain knowledge and preprocessed data, enabling complex querying and visualization of risk factors and their relationships. 
Finally, we construct a BN that leverages both the structured domain knowledge and observations from the data. 
The graphical structure of the BN is derived from the domain knowledge, while its parameters are derived from both the EHR data and domain knowledge represented in the KG. 
Thus, given a multimodal EHR dataset, our approach produces a BN capable of analysing and predicting an individual's disease risk. 
To demonstrate and evaluate the approach, we apply it to a real-world clinical challenge: predicting atrial fibrillation (AF), the most common cardiac arrhythmia, using multimodal EHR data from the Medical Information Mart for Intensive Care (MIMIC)-IV dataset. 

\section{Background and Related Work}

\subsection{Knowledge Graphs and Ontologies}
A KG is a model of real-world domain knowledge using nodes to represent entities of interest and edges to represent the relations between them \cite{hogan_knowledge_2022}. 
This graph structure provides an efficient, intuitive, and integrative abstraction of knowledge, which is particularly useful in the health domain \cite{abu-salih_healthcare_2023}. 
KGs commonly use an ontology-based data schema \cite{abu-salih_healthcare_2023,tamasauskaite_defining_2023}. 
An ontology can be defined as a formal (i.e. machine-readable) and explicit specification of concepts within a domain, based on shared understanding by domain experts \cite{studer_knowledge_1998}. 
Therefore, ontology-based KGs benefit from standardised predefined structure of entities and the relations between them \cite{tamasauskaite_defining_2023}. 
KGs have been widely explored for representing EHR data \cite{murali_towards_2023}, including from the MIMIC dataset \cite{aldughayfiq_capturing_2023}.

\subsection{Bayesian Networks}
Formally, a BN is defined as a pair \begin{math}{\{G, \Theta\}}\end{math}, where \begin{math}{G}\end{math} is a directed acyclic graph and \begin{math}{\Theta}\end{math} is a set of parameters. 
The graph \begin{math}{G}\end{math} consists of nodes that represent variables, which are linked by directed edges that signify direct influences between the linked nodes.
Each node in a BN has a finite set of mutually exclusive states. 
Whereas KGs can have directed or undirected edges, contain cycles, and have multiple edges between the same nodes \cite{hogan_knowledge_2022}, BNs are necessary directed and acyclic, with only one edge between any two nodes. 
Thus two variables \textit{A} and \textit{B} can be represented as \textit{A} \begin{math}{\rightarrow}\end{math} \textit{B}, where \textit{A} is the parent node, \textit{B} is the child node, and \begin{math}{\rightarrow}\end{math} indicates that \textit{A} influences \textit{B}.
The parameters \begin{math}{\Theta}\end{math} determine the strength of the dependencies between the nodes and are defined as conditional probability distributions in a conditional probability table (CPT). 
BNs have been extensively used in the health domain \cite{mclachlan_bayesian_2020} and in health risk prediction in particular \cite{arora_bayesian_2019}.

\subsection{Ontology/KG-Driven BN Construction}
\label{bn-construction}
There has been extensive research on the construction of BNs based on ontologies and KGs.
Fenz \cite{fenz_ontology-based_2012} proposed a method for ontology-based BN construction in which ontology classes and/or individuals are manually selected as BN nodes, after which the BN structure is automatically constructed.
This approach was extended by Ogundele et al. \cite{ogundele_building_2017}, who developed an ontology-driven BN for predicting tuberculosis treatment adherence behaviour based on categorised influencing factors. 
The approach has also been successful in the finance domain for automating share evaluation based on factors that influence the quality and value of individual shares \cite{drake_invest_2022}.

\subsection{Disease Risk Prediction from EHR Data}
Risk scores are often used in clinical practice to systematically stratify patients into different risk levels for certain diseases based on factors such as health history, demographics, and behaviour.
For AF risk prediction, established risk scores include HARMS\textsubscript{2}-AF score \cite{segan_new-onset_2023} and CHARGE-AF \cite{alonso_simple_2013}.
While risk scores are useful, they are limited by their lack of automation and their reliance on complete health records, which may not always be available.

Recent research has explored both KGs and BNs for automated health risk prediction using EHR data. 
Tao et al. \cite{tao_mining_2020} and Chaturvedi et al. \cite{chaturvedi_sequential_2023} developed data-driven KGs for predicting general health risks and type II diabetes, respectively.
However, their lack of an ontology schema limits the explainability afforded by formally represented domain knowledge linking health risks to their influencing factors.
Suo et al. \cite{suo_development_2024} and Bandyopadhyay et al. \cite{bandyopadhyay_data_2015} developed BNs for coronary heart disease and cardiovascular risk, respectively, using domain knowledge for variable identification and EHR data for parameter learning. 
Although these approaches leverage the probabilistic reasoning capabilities of BNs to handle uncertainty, they remain tailored to the datasets they were trained on and lack the semantic interoperability, explainability, and generalisability that ontology-based KGs provide.
Therefore, while existing work has explored data representation and risk prediction from EHR data, no single approach addresses the combined challenges of pragmatic and clinical interoperability \cite{asuncion_pragmatic_2010,benson_principles_2021}, explainability, and uncertainty handling.

In previous work, we developed an ontology-based KG for AF risk prediction with rule-based reasoning and fuzzy inference \cite{nzomo_semantic_achitecture_2024}.
However, this KG was purely knowledge-driven and validated only on simulated data, lacking the population-specific adaptability that data-driven learning provides.
Additionally, while fuzzy inference effectively represents vagueness in clinical parameters, it cannot adequately address missing data.
This paper extends our previous work by incorporating probabilistic reasoning for improved uncertainty handling as well as learning from real-world data.

\section{A Hybrid Approach for Bayesian Network Construction}

\begin{figure*}[!b]
\centerline{\includegraphics[width=\textwidth]{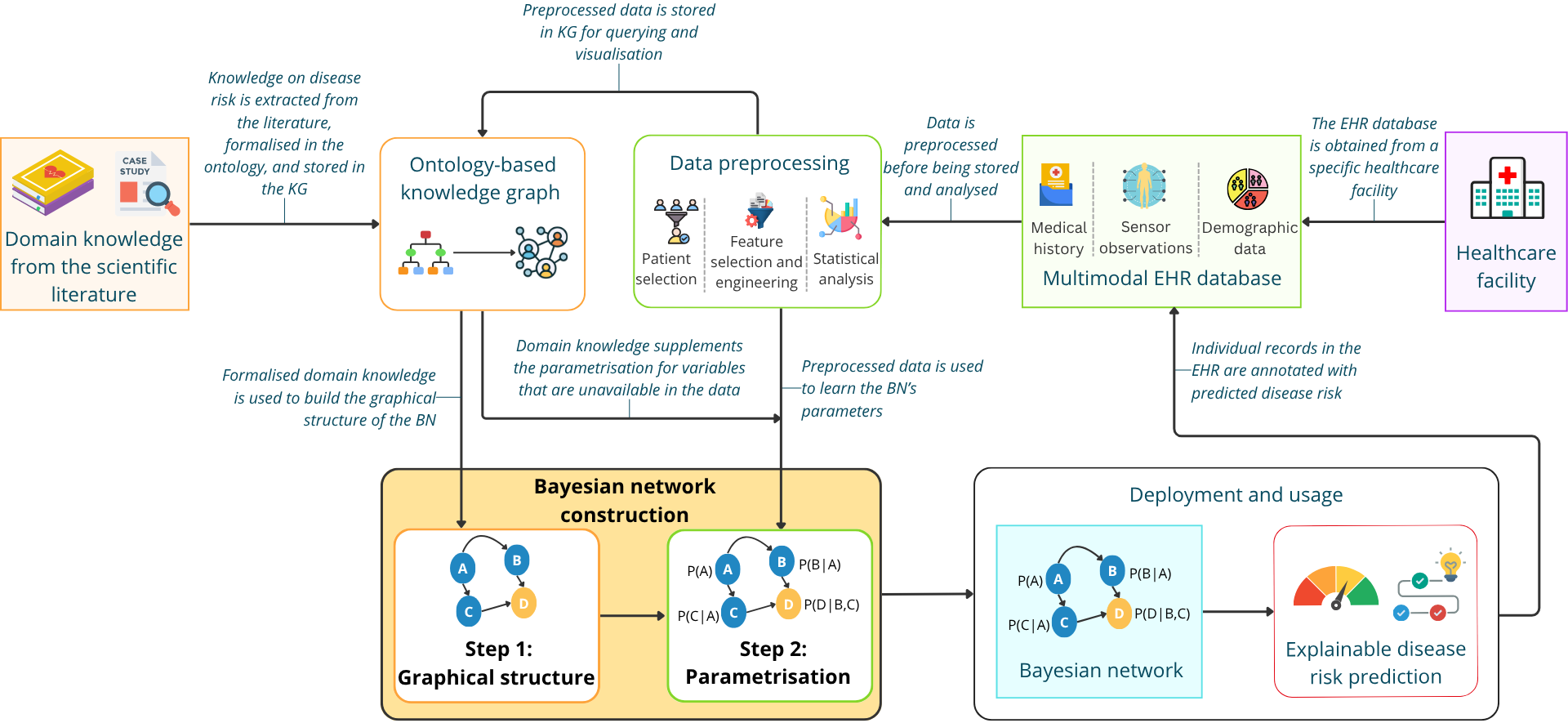}}
\caption{Process flow diagram of the approach for hybrid BN construction for disease risk assessment.}
\label{process}
\end{figure*}

\subsection{Overview of the Approach}
Our approach aims to address the three challenges established in Section~\ref{intro}.
To ensure this, we define five design goals as outlined in Table~\ref{design-goals}. 

\begin{table}[ht]
\centering
\caption{The design goals of the approach and the challenges they address.}
\renewcommand{\arraystretch}{1.5}
\resizebox{\columnwidth}{!}{%
\begin{tabular}{|p{0.05\columnwidth}|p{0.65\columnwidth}|p{0.3\columnwidth}|}
\hline
\textbf{\#} & \textbf{Design Goal} & \textbf{Challenge}\\
\hline
1 & Capture scientific domain knowledge about the factors that influence disease risk. & \makecell[l]{Explainability \\Interoperability}\\
\hline
2 & Process, integrate, and semantically annotate multimodal EHR data, particularly medical history and physiological measurements. & Interoperability \\
\hline
3 & Perform inference while accounting for the non-deterministic nature of health outcomes, even with incomplete information. & Uncertainty handling \\
\hline
4 & Adapt general medical knowledge to specific patient populations and clinical settings. & Interoperability\\
\hline
5 & Ensure transparent and interpretable reasoning for disease risk prediction. & Explainability \\
\hline
\end{tabular}
}
\label{design-goals}
\end{table}

Fig.~\ref{process} illustrates the process flow of the approach.
Using domain knowledge from the scientific literature and multimodal EHR data from a healthcare facility, the approach is used to construct a BN that predicts individualised disease risk under uncertainty while maintaining explainability. 
Each risk prediction is supported by a transparent evidence chain, tracing back through the BN to the underlying domain knowledge via the KG and the healthcare setting via the data.
This allows clinicians to validate risk predictions against scientific evidence and the specific clinical setting.
The EHR can then be annotated with the risk predictions, allowing high-risk individuals to be flagged for clinical follow-up and intervention.

The remainder of this section outlines the methodology of our approach as demonstrated through the AF application use case.
First, we describe the multimodal EHR dataset and its preprocessing for the risk prediction task. 
Next, we detail the development of an ontology that formalises domain knowledge about factors influencing AF risk. 
We then explain how this ontology drives the development of a KG, enabling the visualisation and querying of complex personal data and scientific knowledge. 
Next, we discuss the BN construction process which leverages both the structured domain knowledge and the EHR data, balancing generalised medical knowledge with data-specific context.
Finally, we present a usage scenario, demonstrating how AF risk can be predicted in patients at a healthcare facility.

\subsection{EHR Dataset}
The MIMIC-IV dataset \cite{johnson_mimic-iv_2023} contains EHR data from 299,712 de-identified adult patients admitted to intensive care units and emergency departments at the Beth Israel Deaconess Medical Center in Boston, Massachusetts, between 2008 and 2019. 
The dataset includes diagnoses, demographic information, and anthropometric measurements.
We obtained electrocardiogram (ECG) data from the companion MIMIC-IV-ECG dataset \cite{gow_mimic-iv-ecg_2023}, containing 12-lead ECGs from 161,352 patients sampled at 500 Hz.

\subsubsection{Patient Selection}
27,153 (9.06\%) patients in the MIMIC-IV dataset were diagnosed with AF at some point. 
For inclusion in the predictive dataset, we identified patients who had a documented history of at least three hospital admissions or visits, whereby AF was not diagnosed for at least two of the visits prior to the first AF diagnosis. 
We also ensured that admissions or visits prior to the first AF diagnosis occurred on different dates. 
To prevent data leakage, only data from before the first AF diagnosis date was included. 
This approach let us analyse factors leading to AF diagnosis while ensuring adequate medical history before diagnosis. 

\begin{figure}[ht]
\centerline{\includegraphics[width=\columnwidth]{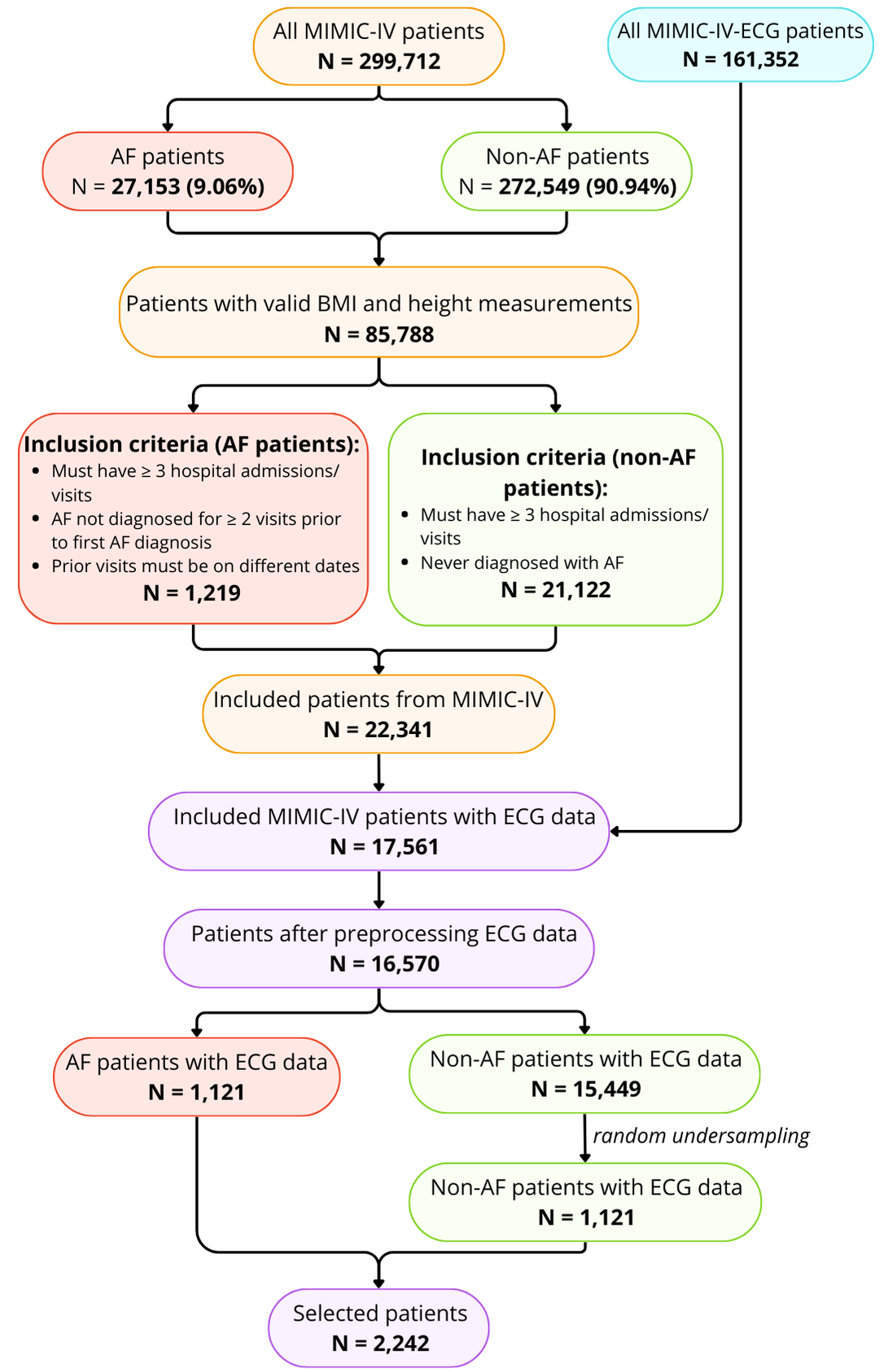}}
\caption{Patient selection process.}
\label{patient-selection}
\end{figure}

We used the same approach (i.e. at least three hospital admissions or visits) to select patients who were never diagnosed with AF. 
This patient selection process was followed by feature selection and engineering as described in Section~\ref{feature-selection}.
The majority class (patients without AF) was then undersampled, resulting in a balanced dataset of 2,242 unique patients.
Fig~\ref{patient-selection} illustrates the inclusion and exclusion of patients at each stage of the patient selection process.

\subsubsection{Risk Factor Selection and Feature Engineering}
\label{feature-selection}
Risk factors were selected based on the scientific literature on AF incidence.
The selected risk factors can be grouped into five categories:

\begin{enumerate}[label=(\roman*)]
    \item Anthropometric factors: These are factors associated with human body measurements, for which we selected BMI and height. 
    Both underweight and obesity increase AF risk compared to normal BMI, with higher obesity levels carrying greater risk \cite{kang_underweight_2016}. 
    For patients with multiple BMI records, we used mean values when measurements were stable (standard deviation $\leq1$) or weighted recent measurements when unstable. 
    We then applied standard adult BMI classifications from underweight to obese class III. 
    Height, which independently increases AF risk \cite{levin_genetics_2020}, was categorised into five sex-specific groups based on population percentiles ranging from $\leq$155 cm to $>$163 cm for females and from $\leq$168 cm to $>$178 cm for males. 
    For patients with multiple height records, we excluded measurements with variations exceeding 2 cm to ensure data quality.
    \item Comorbidity factors: These were assessed through ICD diagnostic codes, which we consolidated into six disease categories: cardiovascular disease, diabetes mellitus, hypertension, kidney disease, chronic obstructive pulmonary disease, and sleep apnoea. 
    Each comorbidity was marked as either present or absent.
    \item Demographic factors: Domain knowledge shows that advancing age, White race, and male sex are linked to increased AF risk \cite{van_gelder_2024_2024}.
    We categorised age into four clinically relevant groups: $<$60, 60-64, 65-74, and $>$74 years.  
    For race, we used MIMIC-IV's race categories Asian, Black, Hispanic, White, and Other, with the last category including inconsistent or missing records. 
    Regarding sex classification, we followed the Health Level Seven (HL7) Gender Harmony model \cite{mcclure_gender_2022} which distinguishes between gender (i.e. gender identity) and sex (i.e. biological sex). 
    We relabelled the MIMIC-IV gender column to reflect its actual reference to sex, as confirmed by the MIMIC-IV documentation\footnote{\url{https://mimic.mit.edu/docs/iv/modules/hosp/patients/}}.
    \item ECG factors: We selected three ECG characteristics that are associated with increased AF risk: prolonged PR interval \cite{aizawa_electrocardiogram_2017}, PR interval variation \cite{chun_electrical_2016}, and abnormal P-wave duration \cite{nielsen_p-wave_2015}. 
    PR interval and P-wave durations in milliseconds (ms) were derived from ECG lead II, which provides a high P-wave strength compared to the other leads \cite{kennedy_detecting_2016}.
    Prolonged PR interval (median $>$200 ms) and PR interval variation (standard deviation $>$ 12 ms) were binary (present or absent), while median P-wave duration was categorised into very short ($\leq$ 89 ms), short (90-99 ms), normal (100-105 ms), intermediate-1 (106-111 ms), intermediate-II (112-119), long (120-129 ms), and very long ($\geq$ 130 ms), based on the study by Nielsen et al. \cite{nielsen_p-wave_2015}. 
    \item Lifestyle factors: Among the behavioural factors linked to increased AF risk, only smoking and alcohol misuse were able to be extracted from the MIMIC-IV dataset through relevant ICD diagnostic codes. Both smoking status and alcohol misuse were simplified to a binary classification (smoker or nonsmoker and present or absent respectively). 
\end{enumerate}

While not exhaustive, the selected risk factors represent the most clinically significant factors that were also available in the MIMIC-IV dataset. 

\subsubsection{Statistical Analysis}
To assess correlations between the selected risk factors and AF diagnosis, we applied Pearson's chi-square test.
Using a contingency table of a factor and the AF status, the test determines association by comparing observed versus expected frequencies \cite{rana_chi-square_2015}. 
P-values below 0.05 indicate significant correlation. 
Our analysis revealed that all the anthropometric, demographic, and ECG factors had zero or near-zero p-values, confirming strong associations with AF. 
We also analysed diagnostic codes (excluding AF codes) present in at least 500 patients. 1,456 codes demonstrated significant association with AF, aligning with domain knowledge on the importance of comorbidities in AF risk assessment. 

P-value does not give an indication of the strength of the association between the factor and AF, merely that an association exists \cite{akoglu_users_2018}. 
We therefore extended our analysis using Cramér’s V, a metric derived from the chi-square statistic to measure the strength of association between categorical variables. 
On scale of 0 to 1, associations can be categorised as weak ($<$ 0.1), moderate (0.1-0.14), strong (0.15-0.24), and very strong ($>$0.24) \cite{akoglu_users_2018}.
Based on these categories, our analysis showed that while some risk factors in the dataset are strongly (race, smoking status, P-wave duration, and prolonged PR interval) and very strongly (age, cardiovascular disease, kidney disease, and PR interval variation) associated with AF, others are only moderately (alcohol misuse, hypertension, and chronic obstructive pulmonary disease) or weakly (BMI, height, sex, diabetes mellitus, and sleep apnoea) associated.

These unexpectedly weaker statistical associations highlight the need for pragmatic interoperability, through which disease risk prediction accounts for specific clinical contexts. 
The unique critical care setting of the MIMIC-IV dataset creates patterns that differ significantly from established epidemiological knowledge. 
Therefore, effective disease risk prediction must balance general domain knowledge with the specific characteristics of the target population.

\subsection{Knowledge Graph Development}
\label{ontology-kg}
The KG was built using a top-down ontology-driven process as defined by Tamašauskaitė and Groth \cite{tamasauskaite_defining_2023}.
We manually developed an ontology to provide a data schema for the KG, specifying the logical constraints of the knowledge to be represented.
In particular, we represent risk factors for AF as identified in Section~\ref{feature-selection}.
Additionally, the ontology was also designed to support the development of both the structure and the parameters of a BN.
This builds upon existing research in ontology-driven BN construction discussed in Section~\ref{bn-construction} \cite{fenz_ontology-based_2012,ogundele_building_2017,drake_invest_2022}, in which influencing factors required for BN reasoning are organised in categories, formalised in an ontology, and assigned a weight value.

To design the ontology, we began by elucidating risk factors from the scientific literature. 
Sources included clinical guidelines on AF, expertly validated AF risk scores, clinical trials and studies, systematic reviews and meta-analyses, and medical textbooks on AF.
We structured the domain knowledge into a conceptual model and implemented it in OWL using Protégé\footnote{\url{https://protege.stanford.edu/}}, an open-source ontology editor and framework.  
We reused several existing ontologies to speed up ontology construction and ensure interoperability with existing resources. 

\begin{figure*}[htbp]
    \begin{subfigure}{\textwidth}
    \centering
    \includegraphics[width=0.85\textwidth]{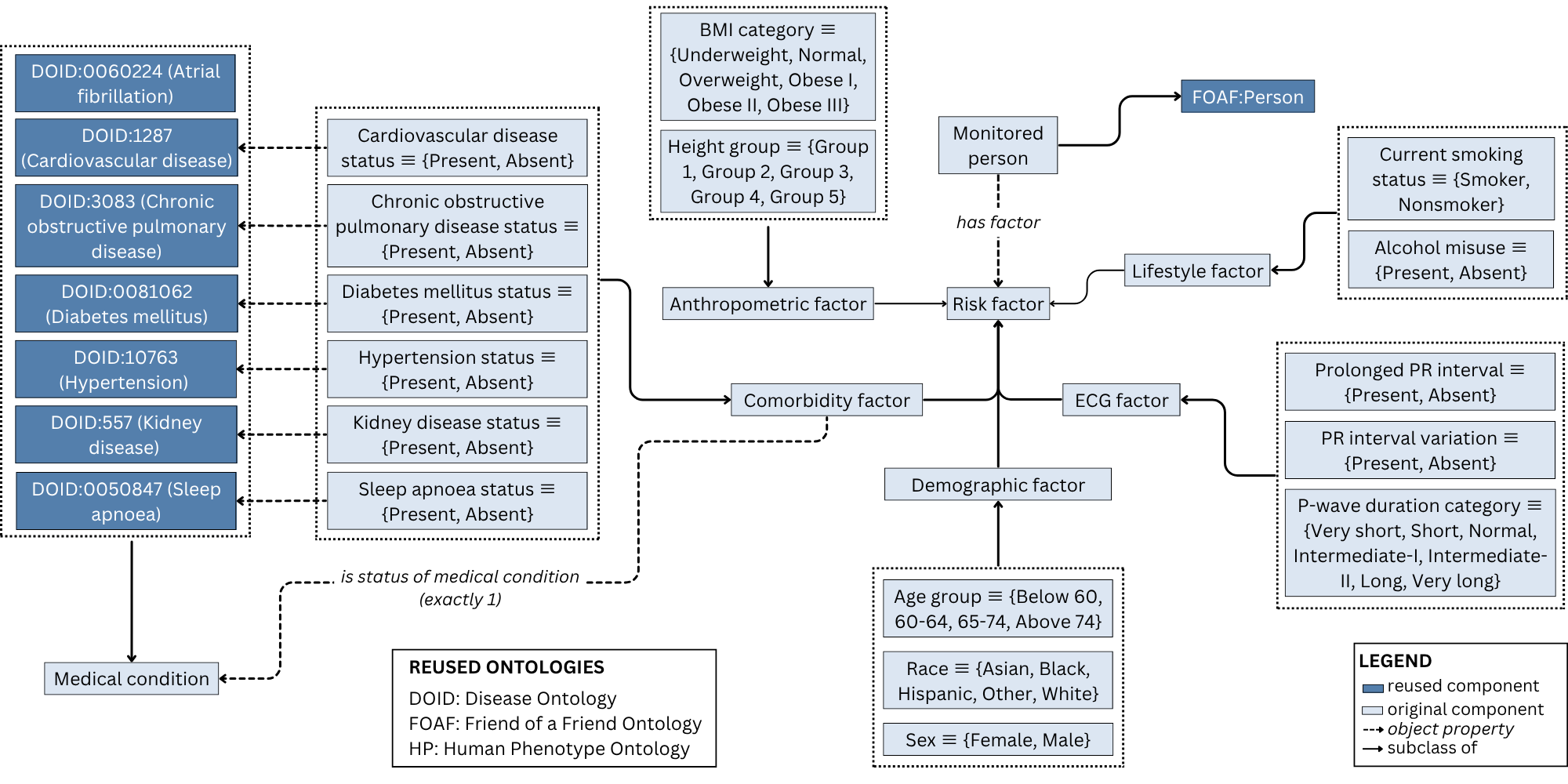}
    \caption{The \textit{Risk factor} class hierarchy.} \label{ontology-a}
    \end{subfigure}   
    \begin{subfigure}{\textwidth}
    \centering
    \includegraphics[width=0.85\textwidth]{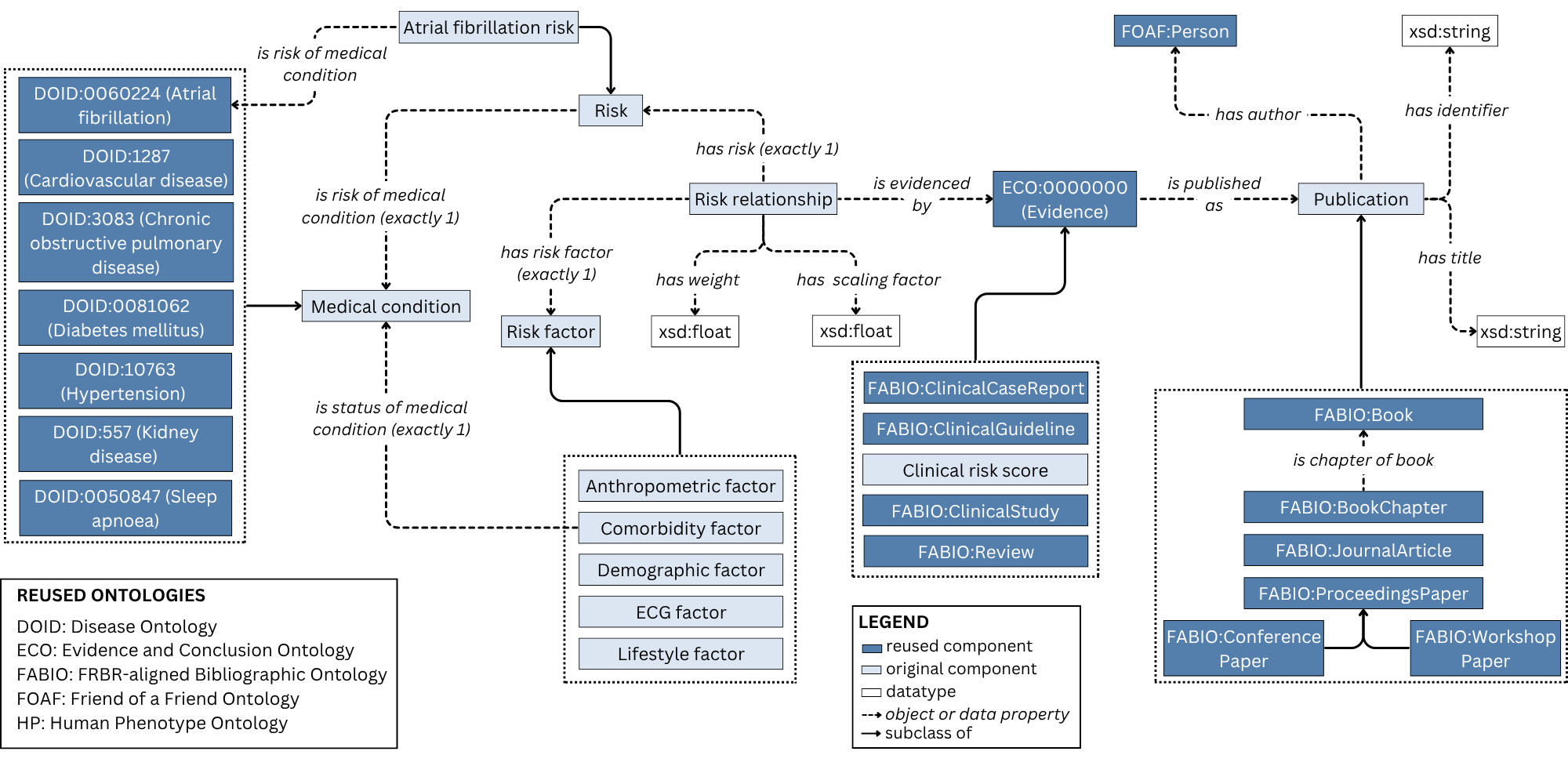}
    \caption{The \textit{Risk relationship}, \textit{Risk factor}, \textit{Evidence}, and \textit{Publication} classes, properties, and relations between them.} \label{ontology-b}
    \end{subfigure}
\caption{Partial visualisation of the ontology.} \label{ontology}
\end{figure*}

We created a \textit{Risk factor} class with five subclasses: \textit{Anthropometric factor}, \textit{Comorbidity factor}, \textit{Demographic factor}, \textit{ECG factor}, and \textit{Lifestyle factor}.  
The \textit{Comorbidity factor} class is linked to a class, \textit{Medical condition}, via an object property, \textit{is status of medical condition}, such that each comorbidity status (present or absent) is linked to its respective medical condition.  
Additionally, each leaf risk factor class is defined by its set of possible values represented as instances.  
For example, the \textit{Age group} class, a subclass of \textit{Demographic factor}, is completely defined by four instances: \textit{Below 60}, \textit{60-64}, \textit{65-74}, and \textit{Above 74}.  
Fig.~\ref{ontology-a} shows the Risk factor class hierarchy.

To model the relationship between risk factor and disease risk, we defined a \textit{Risk relationship} class with two mandatory object properties: \textit{has risk factor}, which links to exactly one \textit{Risk factor} instance, and \textit{has risk}, which links to exactly one \textit{Risk} instance.
\textit{Risk} is defined as a class to represent the risk of a particular condition, and is linked to the \textit{Medical condition} class via an object property, \textit{is risk of medical condition}. 

Evidence supporting each risk relationship is modelled through an object property, \textit{is evidenced by}, which links to the \textit{Evidence} class. 
The \textit{Evidence} class connects to a \textit{Publication} class via the \textit{is published as} property, with the \textit{Publication} class containing properties such as \textit{has title}, \textit{has identifier}, \textit{has date of publication}, and \textit{has author}. 
The \textit{Risk relationship} class also has two important data properties: \textit{has weight}, which represents how much influence a risk factor has on the total risk relative to the other risk factors; and \textit{has scaling factor}, which represents the relative probability of AF for a particular risk factor state. 
This can be based on hazard ratios or risk ratios from scientific studies. 
Fig.~\ref{ontology-b} highlights the \textit{Risk relationship} class and its associated properties and classes.

\begin{figure*}[!t]
\centerline{\includegraphics[width=\textwidth]{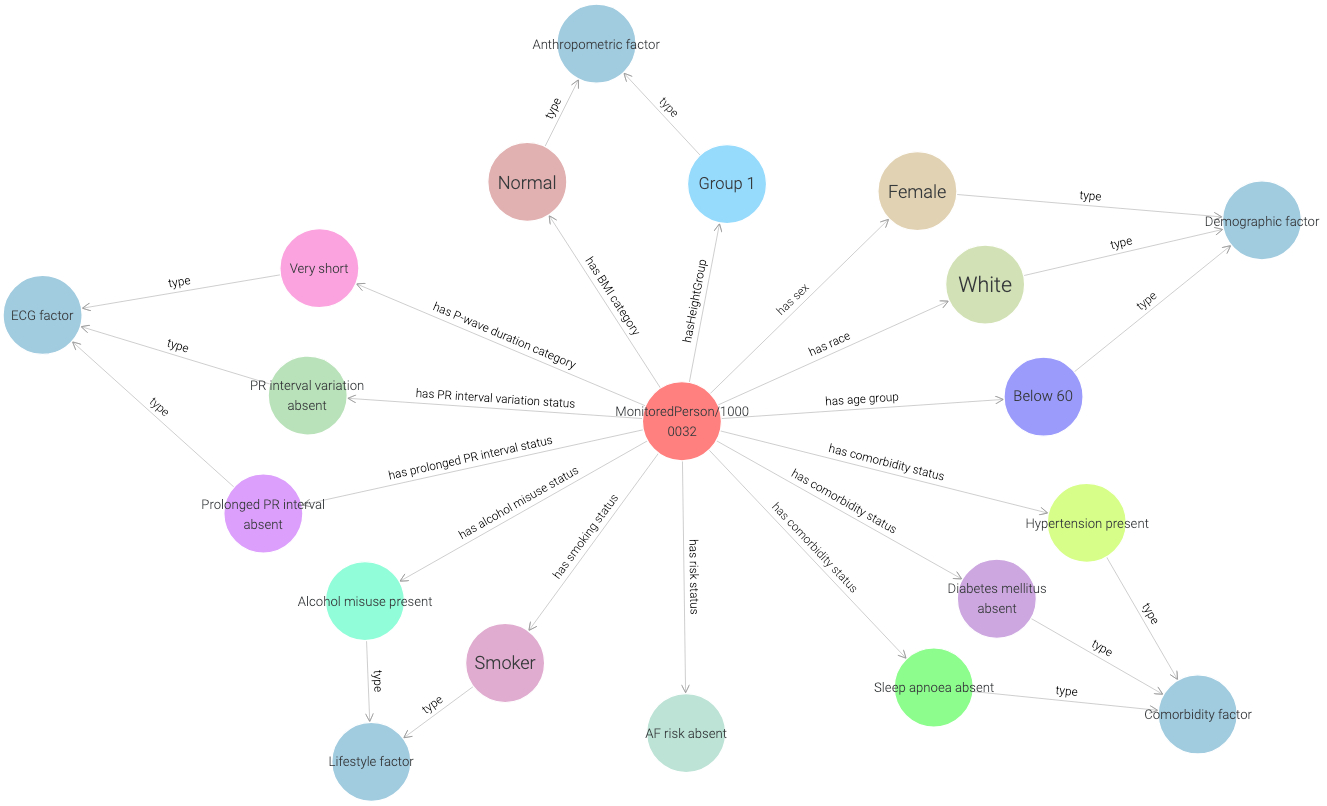}}
\caption{A snippet of the KG in GraphDB showing an individual patient's risk factors and AF risk status.}
\label{kg_snippet}
\end{figure*}

This ontological structure is designed to be extensible.
Additional risk factors and risk relationships, both for AF and other disease, can be added to the ontology, making it suitable for broader clinical applications beyond this specific use case application.
The KG was subsequently constructed using Ontotext GraphDB\footnote{\url{https://www.ontotext.com/products/graphdb/}}, a graph dataset with RDF and SPARQL support that is compliant with W3C standards. 
MIMIC-IV data was manually mapped to the ontology using GraphDB OntoRefine\footnote{\url{https://www.ontotext.com/products/ontotext-refine/}}, a tool that enables the mapping of any structured data to a locally stored RDF schema in GraphDB.
Fig.~\ref{kg_snippet} shows representative snippet of the resulting KG for an individual patient. 
The complete ontology and accompanying documentation are publicly accessible online\footnote{\url{https://mbithenzomo.github.io/afmo/}}.

\subsection{Bayesian Network Development}
\label{bn-development}
There are two integral steps in BN construction: establishing the graphical structure (the different variables and their states) and determining the CPTs to represent the dependencies between variables. These steps can be done in three different ways: using a knowledge-driven approach, a data-driven approach, or a combination of both \cite{mclachlan_bayesian_2020}. Our approach for BN construction is hybrid, in that it aims to combine both domain knowledge and learning from data. To benchmark our approach, we constructed two baseline BNs: one an entirely knowledge-driven BN constructed from the KG, save for the prior probabilities, and the other a purely data-driven BN constructed from the MIMIC-IV dataset. All BNs were constructed using PyAgrum\footnote{\url{https://pyagrum.readthedocs.io}}, a toolbox to build models and algorithms for probabilistic graphical models in Python. 

\subsubsection{Knowledge-driven approach}
The graphical structure of the knowledge-driven BN is based on the risk factors and their corresponding values identified from domain knowledge and represented in the KG. 
We ensured that the risk factor ontology classes could be directly mapped to BN nodes, and their instances mapped to BN states \cite{ogundele_building_2017,drake_invest_2022}.
Each leaf risk factor class becomes a root node in the BN, while the five risk factor category classes are modelled as synthesis nodes following Fenton and Neil’s \cite{fenton_risk_2018} definitional/synthesis idiom.
This idiom allows for intermediate nodes to summarise the effect of a subset of parents on a child. 
Two additional synthesis nodes, \textit{BMI risk} and \textit{Height risk}, were created to calculate the risk of each BMI category and height group respectively, as they differ based on sex. 
Each synthesis node has three states: high, medium, and low. 

\begin{figure*}[!ht]
    \begin{subfigure}{\textwidth}
    \centering
    \includegraphics[width=\textwidth]{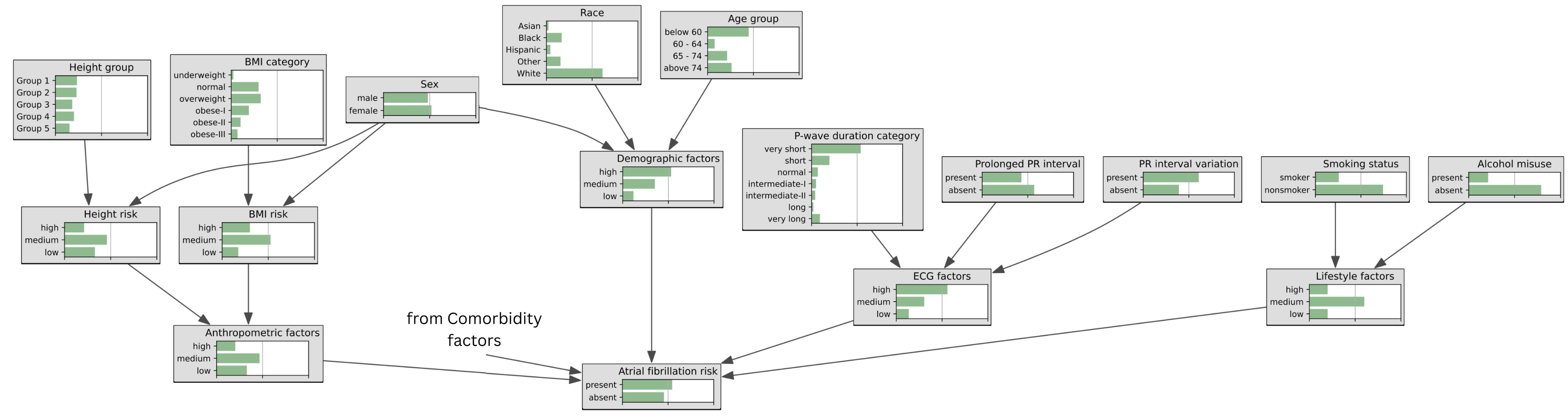}
    \caption{Anthropometric, Demographic, ECG, and Lifestyle factors.} \label{hybrid-bn-a}
    \end{subfigure}   
    \begin{subfigure}{\textwidth}
    \centering
    \includegraphics[width=\textwidth]{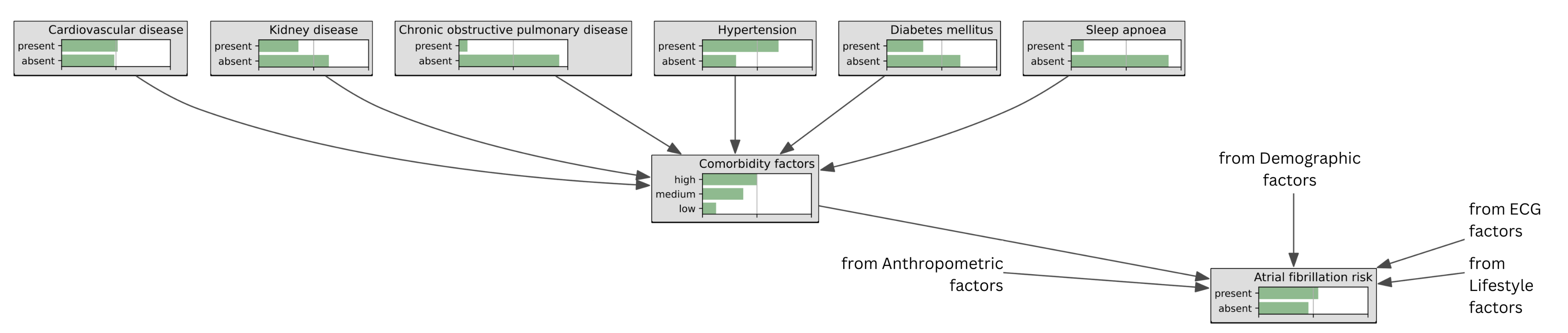}
    \caption{Comorbidity factors.} \label{hybrid-bn-b}
    \end{subfigure}
\caption{Structure and prior probabilities of the hybrid BN.} \label{hybrid-bn}
\end{figure*}

To calculate the CPTs of the synthesis nodes, we defined an algorithm with the following parameters: 

\begin{enumerate}[label=(\roman*)]
    \item Parent node risks: The risk for each parent node was defined as the probability of AF for each state of the node, based on domain knowledge.
    For instance, for the \textit{Sex} node, Segan et al. \cite{segan_new-onset_2023} report a hazard ratio (HR) of 1.84 for males, representing an 84\% increase in AF risk compared to females.
    The domain knowledge-derived risks are represented in the ontology using the \textit{has scaling factor} data property of the \textit{Risk relationship} class.
    All risks were normalised to ensure they summed to 1.
    \item Parent node weights: This represents how much influence a node has in the synthesis node’s total risk relative to the other parent nodes. Thus the weights represent a ratio of each parent node. 
    The weights were manually set based on domain knowledge, represented in the ontology as the \textit{has weight} data property. 
    For instance, for lifestyle factors, the weight ratio of \textit{Smoking status} to \textit{Alcohol misuse} was set as 0.8:0.2, as heavy alcohol consumption has a significantly stronger impact on AF risk than smoking. 
    \item Total risk: The total risk for each synthesis node was then calculated as the sum of each parent node’s risk multiplied by the corresponding weight. 
    \item Thresholds: Two thresholds (T1 and T2) were set to determine the risk state (low, medium, or high) of the synthesis node based on the total risk, such that 0 $<$ T1 $<$ T2 $<$ 1. 
    Risks below T1 result in a low-risk state, those between T1 and T2 result in a medium-risk state, and those higher than T2 result in a high-risk state.
\end{enumerate}

While the knowledge-driven BN's graphical structure and CPTs were knowledge-driven, we learned the prior probabilities of the network from the training data using the MIIC algorithm.
This allowed the distributions of the risk factors to reflect the target population, while keeping the reasoning components knowledge-driven.

\subsubsection{Data-driven approach}
To build the data-driven BN, both structures and parameters were learned from the data.
We evaluated three learning algorithms from PyAgrum's \textit{BNLearner} module: Multivariate Information-based Inductive Causation (MIIC), greedy hill climbing, and local search with tabu list. 
Greedy hill climbing demonstrated marginally better performance and was therefore selected.
We partitioned the dataset into training and testing sets in an 80:20 ratio, with learning restricted to the training set to prevent data leakage. 
The purely data-driven approach optimised for statistical patterns in the data, producing a complex structure with unexpected parent-child relationships and incorrect directional flows between nodes.

\subsubsection{Hybrid approach}
The construction of the graphical structure in the hybrid approach follows the same approach as the knowledge-driven one, with nodes and states derived from risk factor classes and instances in the KG. As with the knowledge-driven BN, prior probability parameters were learned from the MIMIC-IV dataset using the MIIC algorithm. This resulted in identical graphical structures and prior risk factor probabilities between both the knowledge-driven and the proposed models.
The structure of the resultant BN, which we refer to as the hybrid BN, is shown in Fig~\ref{hybrid-bn}.

The hybrid and knowledge-driven BNs differ in how they calculate parent node risks and weights.
For the hybrid BN, we calculated risks directly from the training data by determining the prevalence of AF across each state of each risk factor.
The parent node risks were also obtained from the training data using Cramér’s V to measure the association between the risk factors and AF.
The total risk calculation and threshold-based risk states remained consistent with the knowledge-driven model.
For the synthesis nodes, which are not represented in the data, we used the same algorithm used in the knowledge-driven BN. 
Thus, the hybrid BN construction approach leverages both the structured knowledge representation of the ontology-based KG and the statistical patterns present in the dataset.

\subsection{Usage Scenario}
\label{usage-scenario}
The BN resulting from this approach is designed to run in the back-end of a facility's EHR system, continuously analysing each patient's data, assessing their disease risk, and annotating the EHR with the risk prediction.
This personalised risk prediction, as well as the key contributing risk factors and the scientific evidence supporting the risk relationship, is then made available to clinicians.
Consider a female 78-year-old patient presenting with hypertension, a history of alcohol misuse, and a normal BMI.
Even with this limited information, the BN is able to predict that the individual's AF risk is present with a probability of 63.63\%.
During consultation, the patient and their clinician can view this risk prediction as well as the responsible risk factors.
Each risk factor is accompanied by a brief explanation of its specific relationship to AF risk and its relative contribution within the patient's overall risk profile, with supporting scientific publications fetched from the KG.
Based on this, the clinician can offer personalised approaches to mitigate the risk of AF, for instance by ensuring the patient's hypertension is controlled and referring them for specialised alcohol treatment.

\section{Evaluation and Analysis}

\subsection{Predictive Performance}
\label{prediction}
We compared the hybrid BN with the data-driven and knowledge-driven BNs based on predictive performance using five metrics: accuracy, recall, precision, F1 score, and area under the Receiver Operating Characteristic (ROC) curve (AUC).
We used a threshold of 0.5 for AF risk to be present.
Accuracy represents the percentage of correct risk predictions out of all predictions made by the model.
Recall measures the model's ability to correctly identify patients who are at risk, while precision measures the proportion of patients predicted to be at risk who are actually at risk.
There is usually a trade-off between recall and precision. 
In health risk assessment, high recall is preferable to high precision even at the cost of more false positives, which are acceptable when follow-up tests are accessible and non-invasive.
F1 score provides a harmonic mean between precision and recall, providing a balance between the two in a single metric.
Finally, AUC measures the model's ability to correctly distinguish between high-risk and low-risk patients across different classification thresholds.
The results are shown in Table~\ref{comparison-results}.

\begin{table}[ht]
\centering
\caption{Predictive performance of the three BNs.}
\renewcommand{\arraystretch}{1.5}
\resizebox{\columnwidth}{!}{%
\begin{tabular}{|p{0.3\columnwidth}|p{0.15\columnwidth}|p{0.15\columnwidth}|p{0.15\columnwidth}|p{0.15\columnwidth}|p{0.1\columnwidth}|}
\hline
\diagbox{\textbf{Model}}{\textbf{Metric}} & \textbf{Accuracy} & \textbf{Recall} & \textbf{Precision} & \textbf{F1 Score} & \textbf{AUC}\\
\hline
Knowledge-driven BN & 61.02\% & 41.26\% & 67.65\% & 51.25\% & 0.74\\
\hline
Data-driven BN & \textbf{78.40\%} & 82.96\% & \textbf{75.82\%} & \textbf{79.23\%} & \textbf{0.84}\\
\hline
\textbf{Hybrid BN} & 75.95\% & \textbf{87.00\%} & 71.06\% & 78.23\% & 0.80 \\
\hline
\end{tabular}
}
\label{comparison-results}
\end{table}

The knowledge-driven BN significantly underperformed compared to both the data-driven and hybrid BNs.
It achieved an overall accuracy of 61.02\%, F1 score of 51.25\%, and AUC of 0.74. 
Notably, it had a significantly lower recall (41.26\%) than precision (67.65\%), indicating that it failed to identify a large proportion of high-risk patients.
This shows that the model struggles to distinguish between risk classes in the data, suggesting that a purely knowledge-driven approach is insufficient for risk assessment.

The data-driven BN demonstrated good performance on the testing set, achieving an accuracy of 78.40\%, F1 score of 79.23\%, and AUC of 0.84. 
These metrics indicate reliable discrimination between risk classes. 
The model's precision of 75.82\% is lower than the recall of 82.96\%, suggesting that the data-driven approach is more effective at identifying true positive cases than it is at avoiding false positives. 

The hybrid BN had a highest recall of the three models (87.00\%), demonstrating its strength in identifying true positives. 
In contrast, its precision (71.06\%) was lower than that of the data-driven model.
This suggests that while it is better at capturing true high-risk patients, it also produces more false positives compared to the data-driven BN.
The hybrid BN achieved a more moderate overall performance compared to the data-driven BN, with an accuracy of 75.95\%, F1 score of 78.23\%, and AUC of 0.8. 

\subsection{Scenario Testing}

\begin{figure*}[!t]
  \begin{subfigure}{0.31\textwidth}
  \centering
    \includegraphics[width=\linewidth]{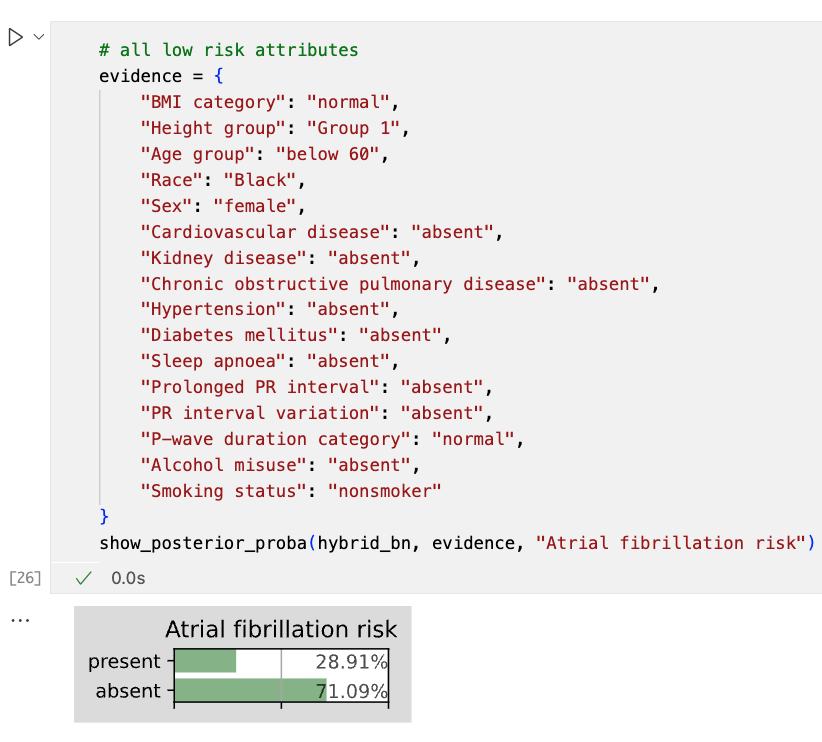}
    \caption{Low-risk scenario} \label{scenario1}
  \end{subfigure}%
  \hspace*{\fill}   
  \begin{subfigure}{0.31\textwidth}
  \centering
    \includegraphics[width=\linewidth]{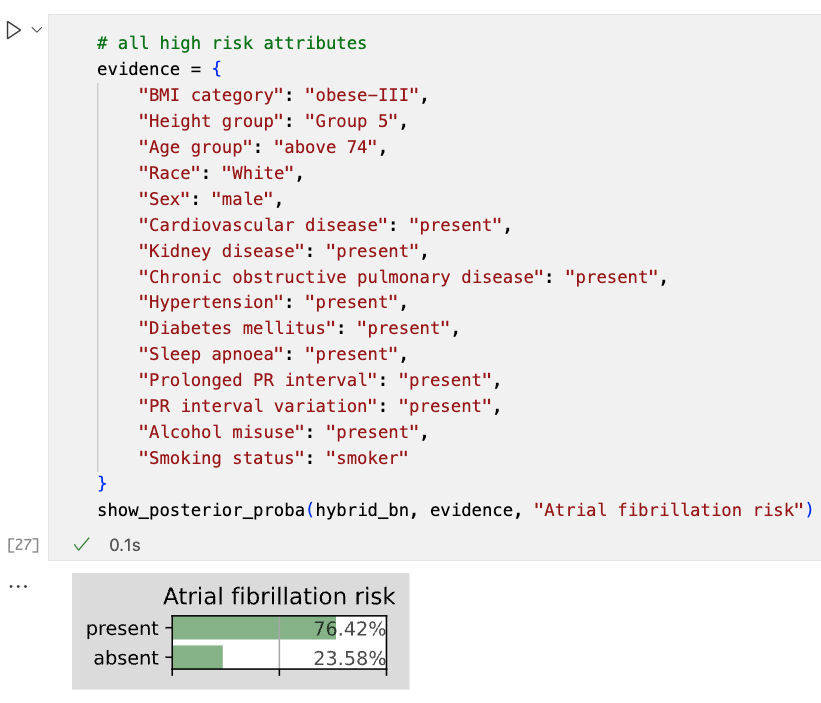}
    \caption{High-risk scenario} \label{scenario2}
  \end{subfigure}%
  \hspace*{\fill}   
  \begin{subfigure}{0.31\textwidth}
  \centering
    \includegraphics[width=\linewidth]{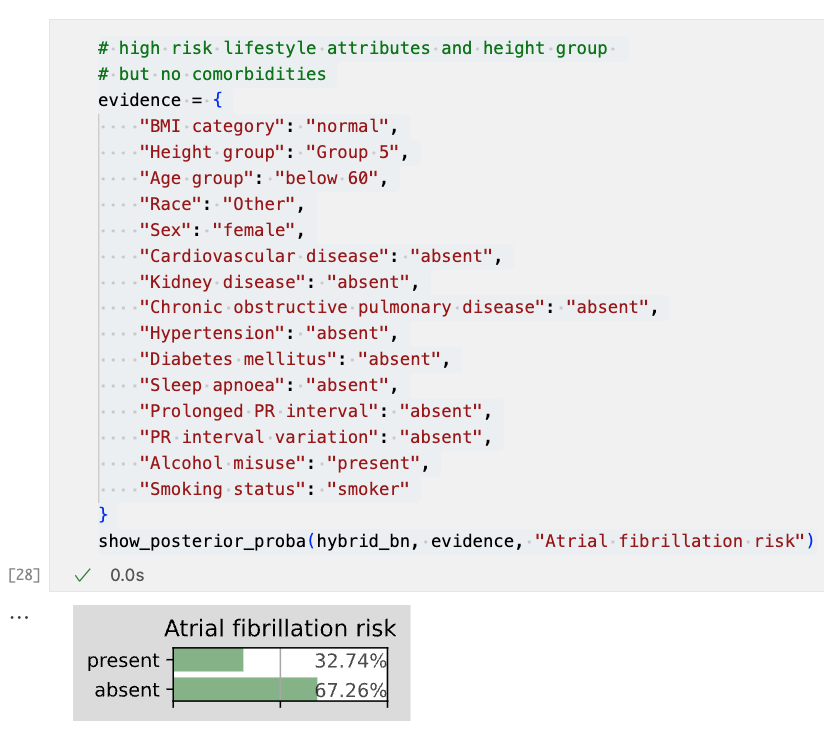}
    \caption{Medium-risk scenario} \label{scenario3}
  \end{subfigure}
  \begin{subfigure}{0.31\textwidth}
  \centering
    \includegraphics[width=\linewidth]{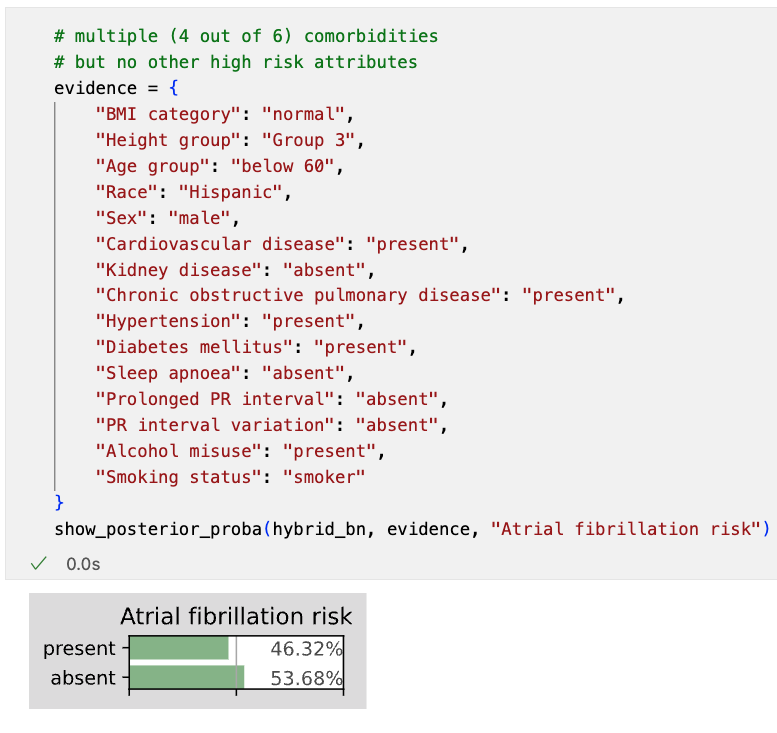}
    \caption{Medium-risk scenario} \label{scenario4}
  \end{subfigure}%
  \hspace*{\fill}   
  \begin{subfigure}{0.31\textwidth}
  \centering
    \includegraphics[width=\linewidth]{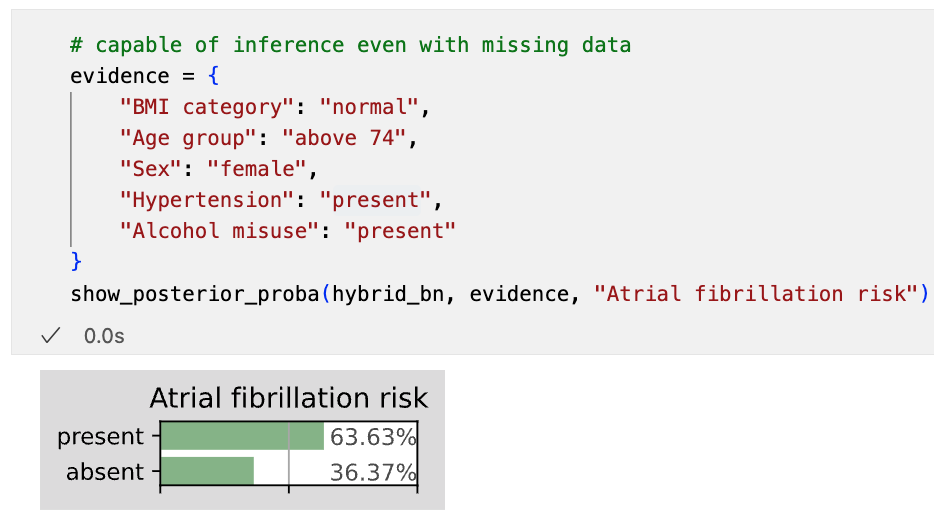}
    \caption{Reasoning with limited evidence.} \label{scenario5}
  \end{subfigure}%
  \hspace*{\fill}   
  \begin{subfigure}{0.31\textwidth}
  \centering
    \includegraphics[width=\linewidth]{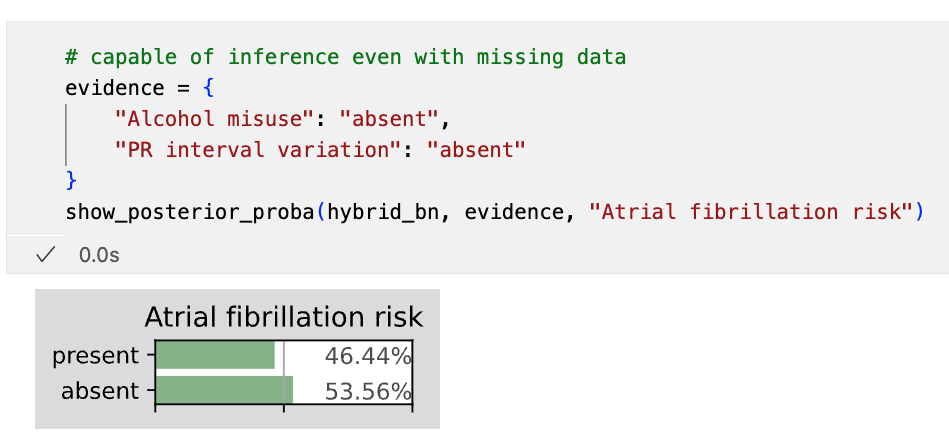}
    \caption{Reasoning with limited evidence.} \label{scenario6}
  \end{subfigure}

\caption{Scenario testing the hybrid BN.} \label{scenarios}
\end{figure*}

We validated the hybrid BN's ability to assess patient risk for AF through scenario testing, simulating six diverse patient profiles including the example detailed in the usage scenario in Section~\ref{usage-scenario}.
We set evidence across different combinations of risk factors to assess the model's responses in low-risk, high-risk, and medium-risk scenarios, as seen in Fig.~\ref{scenarios}.
When simulating low-risk scenarios, such as having a normal BMI and being a nonsmoker, the model appropriately demonstrates low AF risk. 
Conversely, when evidence was set for high-risk attributes such as advanced age and presence of multiple comorbidities, the hybrid BN predicts increased risk. 
The hybrid BN is particularly informative when setting evidence for medium-risk scenarios, where it shows nuanced probability adjustments that reflected the complex interactions between different risk factors. 
Additionally, as seen in Fig.~\ref{scenario5} and Fig.~\ref{scenario6}, the hybrid BN demonstrates the ability to perform inference with limited evidence, reflecting its suitability to real-world settings where data may be incomplete. 
These scenario tests confirm that our hybrid approach not only provides risk assessments that align with clinical knowledge but is also robust to missing data.

\subsection{Interpretability}
\label{interpretability}
While the performance metrics discussed in Section~\ref{prediction} provide valuable insights into the predictive abilities of the models, interpretability is an equally important consideration, particularly in the health domain.
Although the data-driven model achieved higher accuracy, precision, F1, and AUC scores, it prioritised connections between nodes that maximise predictive performance.
This resulted in a graphical structure that was overfitted to the MIMIC-IV dataset and did not align with domain knowledge, limiting its generalisability and utility for clinical decision support.
In contrast, the hybrid BN achieves a good balance between predictive ability and interpretability.
Using the knowledge-driven risk factor hierarchy allows it to capture clinically meaningful relationships between variables that are validated by domain knowledge, making its structure more understandable and supporting explanatory coherence. 
At the same time, its data-driven parameters tailor it to the target population, allowing it to capture the population-specific patterns that differ from general medical knowledge.

\subsection{Achievement of Design Goals}
We can summarise the evaluation of the hybrid approach based on its achievement of the design goals defined in Table~\ref{design-goals}.
The first design goal is achieved through the ontology-based KG, which formalises domain knowledge from the scientific literature as evidence for risk relationships.
The type of evidence and the details of how it is published are explicitly captured in the KG.
We demonstrate how medical history, demographic data, anthropometric measurements, and ECG records can be preprocessed for risk prediction, semantically annotated, and represented in the KG, thus achieving the second design goal.
By incorporating BNs in our approach, we ensure its uncertainty handling capabilities and achieve the third design goal. 
The hybrid BN provides probability distributions rather than definitive predictions, which accounts for non-determinism in health outcomes, while also maintaining reasoning ability when faced incomplete evidence.

The fourth design goal is achieved by combining the knowledge-driven structure with data-driven parametrisation.
This achieves a balance between general domain knowledge and the specific characteristics of the target population derived from observable EHR data.
Finally, the fifth design goal of transparent and interpretable reasoning is also achieved, as discussed in Section~\ref{interpretability}. 
Additionally, as illustrated in the process flow diagram in Fig.~\ref{process}, our approach provides a transparent evidence chain that connects risk predictions to both domain knowledge (from the BN to the KG to the scientific literature) and statistical evidence (from the BN to the data analysis to the raw data).

\section{Discussion and Conclusion}
This paper has presented a novel approach for developing BNs for explainable disease risk prediction, integrating ontology-based KGs with multimodal EHR data.
We demonstrated and validated our approach through an application use case of AF.
The approach achieves pragmatic interoperability, balancing established domain knowledge with adaptability to the specific characteristics of the target population, as validated through our use of the MIMIC-IV critical care dataset.
The resultant BN effectively handles uncertainty by providing probabilistic risk assessments even with incomplete data.
The approach maintains high explainability, with risk predictions able to be traced back to the scientific evidence represented in the KG and the clinical setting from the data.
Further, the combination of graph-based visualisation from the KG and the interpretable graphical structure of the BN provides a clear understanding of the relationships between risk factors and disease risk.

This approach automatically adds personalised risk predictions to patient records, helping identify high-risk individuals for targeted interventions.
The resultant BN demonstrated good performance on the testing set, achieving comparable performance compared to the baseline data-driven and knowledge-driven alternatives.
Future work will refine the approach in several ways.
Firstly, we will enhance the risk stratification by expanding the binary classification into three categories (high, medium, and low risk), thereby improving the granularity of the prediction.
Secondly, we will extend the BN into a Bayesian decision network through the addition of decision and utility nodes, allowing for the recommendation of optimal interventions for at-risk patients.
Thirdly, we intend to implement counterfactual reasoning using the BN, enabling contrastive ``why-not'' explanations to further enhance explainability. 
Finally, we will validate the approach across more diverse and generalised datasets in order to establish its broader applicability.

\bibliographystyle{IEEEtran}
\bibliography{bibliography}

\end{document}